\documentclass{article} 
\usepackage{gpl-iclr2022,times}

\usepackage{amsmath,amsfonts,bm}

\def\eqref#1{equation~\ref{#1}}

\def\1{\bm{1}}

\DeclareMathAlphabet{\mathsfit}{\encodingdefault}{\sfdefault}{m}{sl}
\SetMathAlphabet{\mathsfit}{bold}{\encodingdefault}{\sfdefault}{bx}{n}

\usepackage{hyperref}
\usepackage{url}
\usepackage{multirow}
\usepackage{makecell}

\usepackage{graphicx}
\usepackage{algorithm}
\usepackage{algpseudocode}
\usepackage{booktabs}

\title{Silver-Bullet-3D at ManiSkill 2021:\\Learning-from-Demonstrations and Heuristic Rule-based Methods for Object Manipulation}

\author{Yingwei Pan, Yehao Li, Yiheng Zhang, Qi Cai, Fuchen Long, Zhaofan Qiu, Ting Yao, Tao Mei\\
	JD Explore Academy, Beijing, China\\
	\texttt{\{panyw.ustc,yehaoli.sysu,yihengzhang.chn,cqcaiqi\}@gmail.com}\\
    \texttt{\{longfc.ustc,zhaofanqiu,tingyao.ustc\}@gmail.com, tmei@jd.com}
}

\iclrfinalcopy 
\begin{document}	
	\maketitle
	\begin{abstract} 
		This paper presents an overview and comparative analysis of our systems designed for the following two tracks in SAPIEN ManiSkill Challenge 2021 \footnote{\url{https://sapien.ucsd.edu/challenges/maniskill2021/}}:
		
		\textbf{No Interaction Track}: The No Interaction track targets for learning policies from pre-collected demonstration trajectories. We investigate both imitation learning-based approach, \emph{i.e.}, imitating the observed behavior using classical supervised learning techniques, and offline reinforcement learning-based approaches, for this track. Moreover, the geometry and texture structures of objects and robotic arms are exploited via Transformer-based networks to facilitate imitation learning.
		
		\textbf{No Restriction Track}: In this track, we design a Heuristic Rule-based Method (HRM) to trigger high-quality object manipulation by decomposing the task into a series of sub-tasks. For each sub-task, the simple rule-based controlling strategies are adopted to predict actions that can be applied to robotic arms.
		
		To ease the implementations of our systems, all the source codes and pre-trained models are available at \url{https://github.com/caiqi/Silver-Bullet-3D/}.
	\end{abstract}
	
	\section{Introduction} 
	It is not trivial to learn generalized policies for object manipulation from 3D visual inputs. The difficulty arises from the complexity of controlling robotic arms of multi-degree-of-freedom (multi-DOF) and the diversity of 3D shapes. Most previous benchmarks are insufficient
	for learning generalizable manipulation skills. On the one hand, many existing manipulation tasks \cite{kolve2017ai2,wu2018building,puig2018virtualhome,xia2018gibson,xia2020interactive} only support high-level planning with abstract action space and are thus unsuitable for studying more challenging scenarios which require low-level full-physics simulation. On the other hand, most benchmarks \cite{zhu2020robosuite,urakami2019doorgym,yu2020meta,james2020rlbench,dosovitskiy2017carla,xu2019deep} lack either task-level diversity or object-level variances. Compared with previous benchmarks, ManiSkill \cite{mu2021maniskill} is much more challenging but better resembles common real-world robotics setups. In this work, we aim at exploiting imitation learning and offline reinforcement learning to learn generalizable manipulation skills from pre-collected demonstrations. In addition, we also explore the possibility of interacting with the simulation environments and design a Heuristic Rule-based Method (HRM).
	
	\section{Learning-from-Demonstrations in No Interaction Track}
	\subsection{System}
\begin{figure}[!tb]
		\centering
		\includegraphics[width=0.98\linewidth]{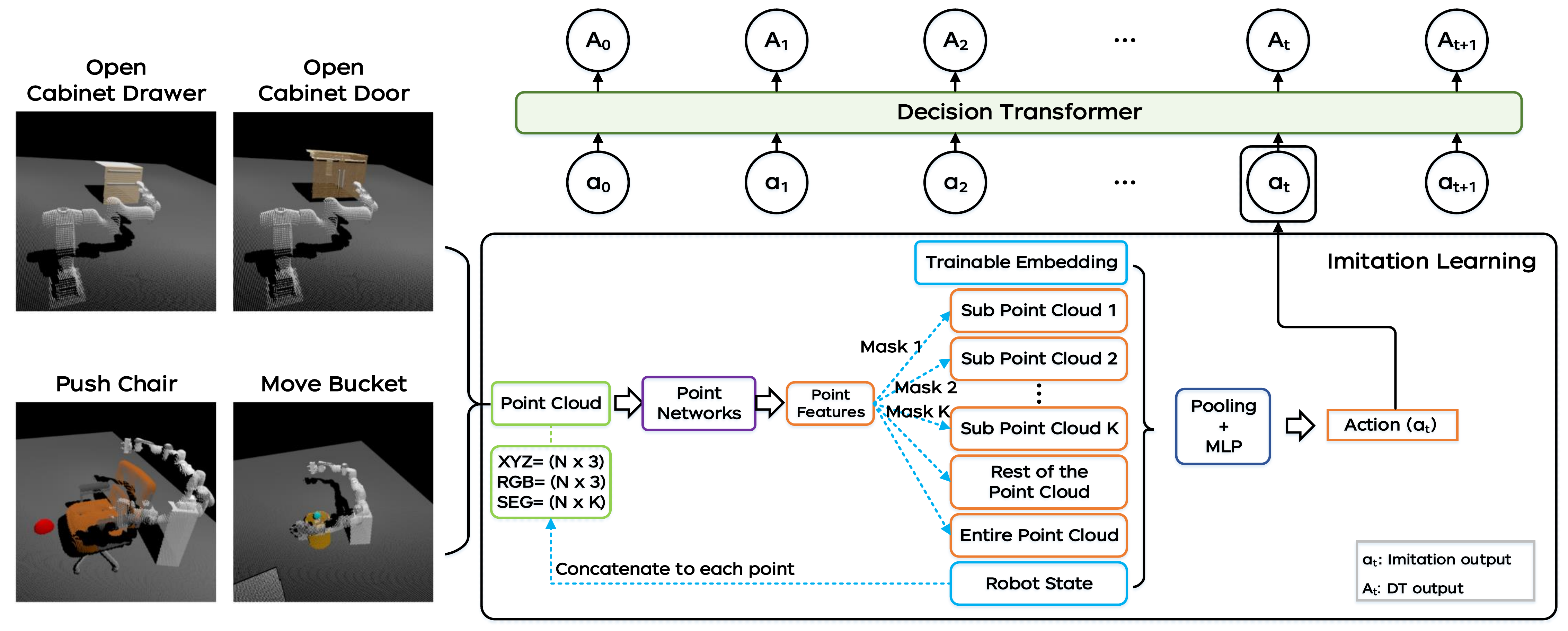}
		\caption{An overview of our solution for No Interaction Track by exploiting imitation learning and Decision Transformer.}
		\label{fig:il_dt_model}
	\end{figure}

	For No Interaction Track, we design our system in two directions: imitation learning and offline reinforcement learning. In imitation learning, the policy is trained with supervised learning that directly imitates the behaviors of provided demonstrations. More specifically, give the observation ($s$) - action ($a$) pairs as training samples, the imitation learning optimizes the function by approximating $f_\theta (s)\xrightarrow[]{}a$. Unlike imitation learning which highly hinges on the quality of the data-collecting process, offline reinforcement learning is more tolerant to the noise of demonstrations. Therefore, inspired by Transformer-based sequence modeling in vision tasks \cite{li2021contextual,luo2021coco,li2021scheduled,pan2020auto,pan2020x,li2022cvpr,Long:CVPR22}, we leverage Decision Transformer (DT) \cite{chen2021decision} to cast the task of manipulation objects as a sequence modeling problem. Figure \ref{fig:il_dt_model} illustrates the overview of our Imitation Learning and Decision Transformer architecture (namely IL-DT), which will be described in the following sections.
	
	\subsubsection{Imitation Learning} Each demonstration consists of a sequence of observations $s=(p, c, m, g)$ and actions $a$, where $p$ is the observed point cloud and $c$ is the color of point cloud. $m$ denotes the mask for point cloud, which groups the point cloud into several parts like robotic arms, cabinet doors, etc. $g$ represents the state of robotic arms. To model the geometric information of objects and robotic agents, we calculate the relative distance between different parts, which is regarded as geometric feature. More specifically, by denoting the finger position and center position of the robotic agent as $g_f$ and $g_c$ respectively, the geometric feature is measured as:
	\begin{equation} \label{geometry}
		x = concat(p, p - g_f, p - g_c).
	\end{equation}
	Next, a three-stage network is designed to learn representative features to manipulate objects.
	
	\textbf{Point-level features.} The feature of each point is composed by two parts: geometric information and texture information. Specifically, the geometric information defined in Eq. (\ref{geometry}) is fed into an MLP network and the RGB information $c$ is fed into another MLP network to extract features. The state of robotic arms $g$ is further embedded into dense robotic features. The geometric features, texture features, and robotic features are concatenated as the point-level representation for each point.
	
	\textbf{Part-level features.} Since different parts of the manipulated objects and robotic arms contain diverse topological and geometric properties, we then group the points according to the provided mask $m$. Each group of points is processed by a separate MLP network to learn each part's unique features. After that, we perform mean pooling over the features of points belonging to the same part, yielding the part-level features.
	
	\textbf{Task-level features.} To explore the relationships among different parts, we further employ a Transformer \cite{vaswani2017attention} network over the part-level features. Specifically, the features of different parts are concatenated as a sequence, which are augmented with the self-attention mechanisms via Transformer. Finally, by concatenating the features of all parts, we feed them into a MLP network for action prediction.
	
	To optimize the network, we combine the L1 loss and L2 loss between prediction $\hat{a}$ and target $a$ as the objective:
	\begin{equation}\label{loss:imitation}
		l = ||\hat{a} - a||_2 + \beta  || \hat{a} - a||,
	\end{equation}
	where $\beta$ denotes the loss weight that controls each component's contribution.

	\subsubsection{Decision Transformer} Vallina imitation learning commonly treats each observation pair independently, leaving the temporal information within the sequence of actions unexploited. Therefore, we further design a Transformer-based network to explore the dependency over timestamps. Specifically, inspired by Decision Transformer (DT) \cite{chen2021decision}, we take the predictions $(\hat{a}_i, \hat{a}_2, ..., \hat{a}_T)$ from imitation learning network as the sequence input, where $T$ is the maximum timestamp. A Transformer network absorbs the sequence and predicts refined action for each timestamp. Following \cite{chen2021decision}, we replace the summation/softmax over all tokens with only the previous tokens in the sequence to enable auto-regressive prediction. We leverage the pre-trained imitation learning networks to initialize the weights of DT-based networks. The overall loss is the same as Eq. (\ref{loss:imitation}) in imitation learning.
	
	\subsubsection{Ensemble Models}
	To improve the robustness of predicted actions, we ensemble multiple models including Imitation Learning-based models and Decision Transformer-based models. We use a late fusion strategy where the actions of all models are averaged as the final prediction. Experimental results demonstrate that the ensemble models exhibit better performances than single models (see Sec. \ref{sec:exp}).
	
	\subsection{Experimental Results}\label{sec:exp}
	
	In this section, we compare the performances of Imitation Learning-based model, Decision Transformer-based model, and Ensemble Models for four tasks in the No Interaction track. Table \ref{tab:no_inter_com} details the performance comparisons in between. In general, the results across different tasks consistently validate the complementarity between Imitation Learning-based model and Decision Transformer-based model.
	
	\begin{table*}[ht]
		\setlength\tabcolsep{3pt}
		\caption{\small The performances of four manipulation tasks on No Interaction track.}
		\centering
		\begin{tabular}{c|cccc}
			\hline
			Method               & \scriptsize{OpenCabinetDoor} & \scriptsize{OpenCabinetDrawer} & \scriptsize{PushChair} & \scriptsize{MoveBucket} \\ \hline
			Imitation-Learning   & 0.880          & 0.890      &  0.397  &   0.663   \\
			Decision-Transformer & 0.893          & 0.893      &  0.413  &  0.670    \\
			Ensemble Models      & 0.900          & 0.917      &  0.433  &  0.687  \\
			\hline
		\end{tabular} \label{tab:no_inter_com}
	\end{table*}
	Table \ref{tab:track1_leaderboard} further illustrates the final performances of our submission on the leaderboard for four different manipulation tasks. The best single model achieves an average score of 0.533, which has achieved better performances than other teams. Next, by consolidating the ensembling of Imitaion Learning and Decision Transformer-based models, our ensemble system leads to 3.4\% absolute improvement against the single run on the first stage evaluation. On the second stage evaluation, our ensemble model achieves the best performance (57.4\%), surpassing the second team by 16.7\% absolute improvement.
	
	\begin{table*}[ht]\small
		\setlength\tabcolsep{1.pt}
		\caption{\small Comparisons of different teams for No Interaction task on leaderboard. Results are directly taken from \url{https://sapien.ucsd.edu/challenges/maniskill2021/result}.}
		\begin{tabular}{c|c|ccccccccc}
			\Xhline{2\arrayrulewidth}
			\makecell{Evaluation \\ Stage} & Team  & \makecell{\scriptsize{MoveBucket} \\ \scriptsize{Train}} & \makecell{\scriptsize{OpenDoor} \\ \scriptsize{Train}} & \makecell{\scriptsize{OpenDrawer}  \\ \scriptsize{Train}} & \makecell{\scriptsize{PushChair} \\ \scriptsize{Train}} & \makecell{\scriptsize{MoveBucket} \\ \scriptsize{Test}} & \makecell{\scriptsize{OpenDoor} \\ \scriptsize{Test}} & \makecell{\scriptsize{OpenDrawer} \\ \scriptsize{Test}} & \makecell{\scriptsize{PushChair} \\ \scriptsize{Test}} & \makecell{\scriptsize{Final} \\ \scriptsize{Score}}                   \\ \hline
			& \makecell{Silver-Bullet-3D \\ (Single)}
			& 0.632      & 0.920      & 0.968      & 0.408      & 0.488      & 0.176      & 0.280      & 0.392      & 0.533          \\
			\multirow{-2}{*}{First}  & \makecell{Silver-Bullet-3D \\ (Ensemble)}
			& 0.744      & 0.944     & 0.968       & 0.456      & 0.528      & 0.168      & 0.400      & 0.328      & 0.567           \\ \hline\hline
			& \textbf{Silver-Bullet-3D}   & \textbf{0.716}  & \textbf{0.896}  & \textbf{0.932}  & \textbf{0.468}  & \textbf{0.488}  & \textbf{0.208}  & \textbf{0.556}  & \textbf{0.328}  & \textbf{0.574}  \\
			& Fattonny           & 0.320  & 0.764  & 0.840  & 0.400  & 0.336  & 0.160  & 0.184  & 0.252  & 0.407  \\
			& bigfish            & 0.360  & 0.664  & 0.740  & 0.256  & 0.260  & 0.124  & 0.192  & 0.152  & 0.344  \\
			& MI                 & 0.268  & 0.700  & 0.788  & 0.320  & 0.156  & 0.124  & 0.120  & 0.180  & 0.332  \\
			& SieRra11799        & 0.196  & 0.200  & 0.456  & 0.212  & 0.144  & 0.044  & 0.144  & 0.120  & 0.190  \\
			& ic                 & 0.108  & 0.280  & 0.484  & 0.156  & 0.060  & 0.072  & 0.228  & 0.116  & 0.188  \\
			\multirow{-7}{*}{Second}
			& Zhihao             & 0.176  & 0.272  & 0.416  & 0.216  & 0.104  & 0.068  & 0.120  & 0.096  & 0.184 \\ \Xhline{2\arrayrulewidth}
		\end{tabular} \label{tab:track1_leaderboard}
	\end{table*}
	
	\section{Heuristic Rule-based Method in No Restriction Track} 
	
	\subsection{System}
\begin{figure}[!tb]
		\centering
		\includegraphics[width=0.98\linewidth]{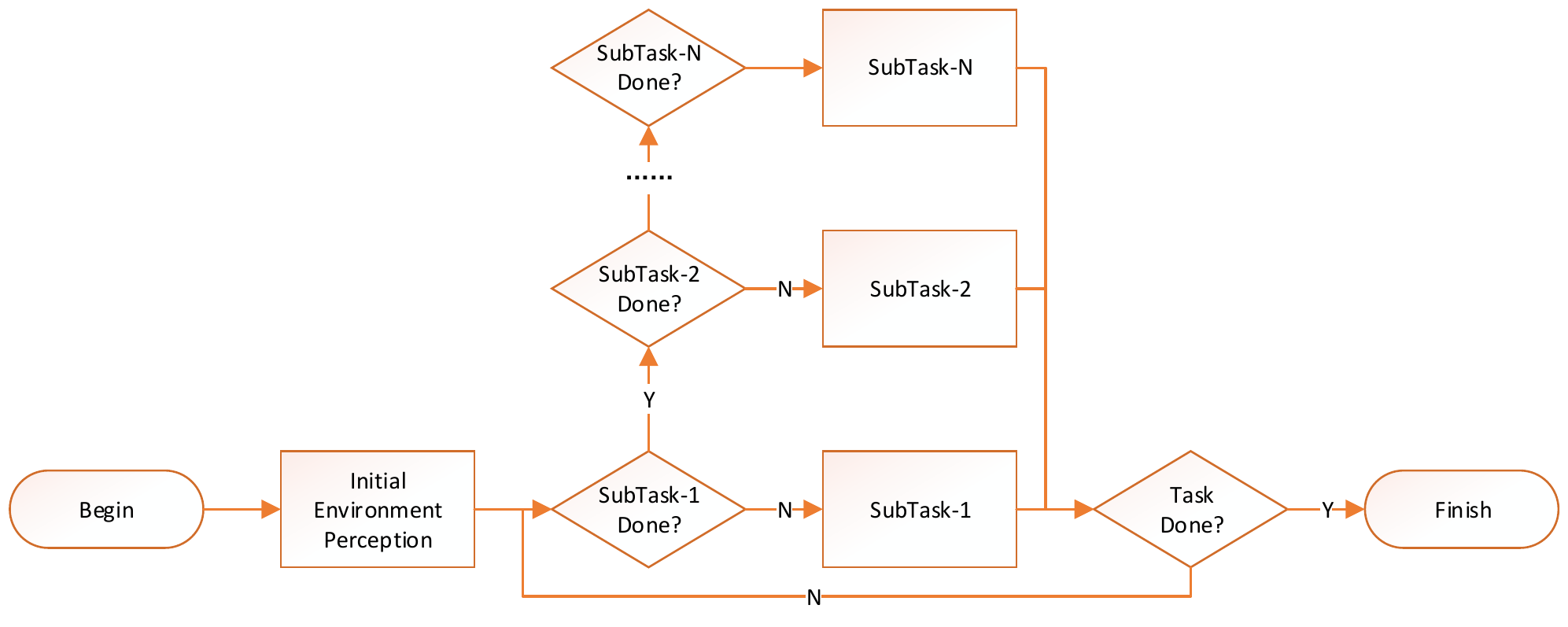}
		\caption{An overview of our Heuristic Rule-based Method (HRM).}
		\label{fig:rule-framework}
	\end{figure}

	The basic idea of our Heuristic Rule-based Method (HRM) is to facilitate object manipulation by decomposing the task into a series of sub-tasks. Here each sub-task could be accomplished with simple rule-based controlling strategies.
	Figure \ref{fig:rule-framework} illustrates an overview of our HRM. At the beginning of a manipulation task, HRM first estimates the attributes (e.g. location, shape, size, and orientation) of the robot, target object and target point (for MoveBucket and PushChair) from the observations. This initialization process is denoted as ``Environment Observation" in Figure \ref{fig:rule-framework}. Next, in each ``observation-action" iteration, the first unfinished sub-task is carried out for one step. Such iterations will stop when the robot successfully manipulates the object or takes more than 200 steps.
	
	Based on the workflow described in Figure \ref{fig:rule-framework}, the central core in HRM is the design of sub-tasks for each type of manipulation. Each sub-task mainly consists of two components: observation and action. Here we introduce the detailed algorithms adopted in several kinds of sub-tasks.
	
	\textbf{Sub-Task: MoveSteps} (Algorithm \ref{alg:sub1}).
	This is a simple but widely utilized sub-task. In this sub-task, the robot is designed to take the same action in a pre-determined number of steps.
	\begin{algorithm}[]
		\caption{Sub-Task: MoveSteps.}\label{alg:sub1}
		\begin{algorithmic}[1]
			\State \textbf{Input:} Fixed Action Vector $A$, number of steps $N$.
			\State \textbf{Output:} Action Vector $\alpha$
			\State \textbf{Initialization:} $i=0$, done=False
			\While{not done}
			\State $\alpha=A$
			\State \textbf{Update:} $i=i+1$, done=$i \geq N$
			\EndWhile
		\end{algorithmic}
	\end{algorithm}
	
	\textbf{Sub-Task: MoveTo} (Algorithm \ref{alg:sub2}).
	This is another basic sub-task. Here the robot is enforced to take the same action until reaches the target location.
	To avoid the disturbance between different moving directions, only one direction in the action vector is activated in one sub-task. During the moving, the location of the robot could be obtained from the \textit{state} in observation, and the target location is determined according to the object or target point. The error threshold ensures that the sub-task could be finished in a limited number of steps.
	\begin{algorithm}[]
		\caption{Sub-Task: MoveTo.}\label{alg:sub2}
		\begin{algorithmic}[1]
			\State \textbf{Input:} Index of activated direction in action vector $i$, target location $x_t$, observation $obs$, velocity $v$, error threshold $t$.
			\State \textbf{Output:} Action Vector $\alpha$
			\State \textbf{Initialization:} Basic action vector $A_0=[0, 0,\cdots, 0]$, done=False
			\While{not done}
			\State $x=process\_obs(obs)$ \Comment{Get current location from observation.}
			\State $d=x_t-x$ \Comment{Calculate distance between target and current location.}
			\State $\alpha=A_0$
			\If{$d > 0$}
			\State $\alpha[i]=v$
			\Else
			\State $\alpha[i]=-v$
			\EndIf
			\State \textbf{Update:} done=$|d| < t$ \Comment{Keep moving until the distance is smaller than the threshold.}
			\EndWhile
		\end{algorithmic}
	\end{algorithm}
	
	\textbf{Arm Stabilization}.
	The pose of the arms is heavily influenced by the reaction force from objects. To keep the pose of arms stable during manipulation, we design a degenerative stabilizer for arm stabilization. The stabilizer is initialized with the reference pose of the arm. Then a ``MoveTo'' sub-task whose target is the reference pose is associated with the arm. The error threshold is set to 0.01. The output of the stabilizer is added to the action vector produced by main-stream sub-tasks.
	
	\textbf{Solutions.}
	Based on such basic designs, we can represent the solution of the manipulation tasks as ordered lists of sub-tasks. Here we summarize the solutions for \textit{OpenCabinetDoor}, \textit{OpenCabinetDrawer}, \textit{MoveBucket}, and \textit{PushChair}.
	
	\textit{OpenCabinetDoor}
	\begin{itemize}
    \item Initialize the pose of arm (MoveSteps subtask);
    \item Rotate the moving platform of the robot till the robot faces the cabinet (MoveTo subtask);
    \item Adjust the height of the moving platform till the fingers have the same height as the handle of the door (MoveTo sub-task);
    \item Adjust the location along Y-axis of the moving platform till the fingers have the same Y as the handle of the door (MoveTo sub-task);
    \item Adjust the location along X-axis of the moving platform till the fingers have the same X as the handle of the door (MoveTo sub-task);
    \item Grasp the handle of the door (MoveSteps sub-task);
    \item Open the door (MoveSteps sub-task).
	\end{itemize}
	
	\textit{OpenCabinetDrawer}
	\begin{itemize}
		\item Initialize the pose of arm (MoveSteps sub-task);
		\item Rotate the moving platform of the robot till the robot faces the cabinet (MoveTo sub-task);
		\item Adjust the height of the moving platform till the fingers have the same height as the handle of the drawer (MoveTo sub-task);
		\item Adjust the location along Y-axis of the moving platform till the fingers have the same Y as the handle of the drawer (MoveTo sub-task);
		\item Adjust the location along X-axis of the moving platform till the fingers have the same X as the handle of the drawer (MoveTo sub-task);
		\item Grasp the handle of the drawer (MoveSteps sub-task);
		\item Open the drawer (MoveSteps sub-task).
	\end{itemize}
	
	\textit{MoveBucket}
	\begin{itemize}
		\item Initialize the pose of arms (MoveSteps sub-task);
		\item Rotate the moving platform of the robot till the robot faces the bucket (MoveTo sub-task);
		\item Hold the bucket with two arms (MoveSteps sub-task);
		\item Initialize the degenerative stabilizer for arm stabilization;
		\item Lift up the bucket with two arms (MoveSteps sub-task);
		\item Rotate the moving platform of the robot till the robot faces the target platform (MoveTo sub-task);
		\item Adjust the location along X- and Y-axis of the moving platform till the robot reaches the edge of the target platform (MoveTo sub-task);
		\item Put down the bucket (MoveSteps sub-task).
	\end{itemize}
	
	\textit{PushChair}
	\begin{itemize}
		\item Move forward along X-axis for several steps to get closed to the chair (MoveSteps sub-task);
		\item Adjust the height of the moving platform till the arms have the same height as the armrest of the chair (MoveTo sub-task);
		\item Hold the chair with two arms (MoveSteps sub-task);
		\item Initialize the degenerative stabilizer for arm stabilization;
		\item Rotate the moving platform till the robot faces the target red point (MoveTo sub-task);
		\item Adjust the location along X- and Y-axis of the moving platform till the chair reaches the target red point (MoveTo sub-task).
	\end{itemize}
	
	\subsection{Experimental Results}
	
	Table \ref{tab:no_restriction} illustrates the final performances of our submission on the leaderboard in No Restriction track. By consolidating the divide-and-conquer strategy which decomposes the complex task into several consecutive simple sub-tasks, we achieve a 0.928 final score, which is 23.8\% absolute higher than second place. Moreover, the performances on the train set and test set are greatly reduced due to the non-data-dependent characteristic of our designed method.
	
	\begin{table}[ht]
		\small
		\centering
		\setlength\tabcolsep{1.5pt}
		\caption{\small  Comparisons of different teams for No Restriction task on leaderboard. Results are directly taken from \url{https://sapien.ucsd.edu/challenges/maniskill2021/result}.}
		\begin{tabular}{c|ccccccccc}\toprule
			
			Team  & \makecell{\scriptsize{MoveBucket} \\ \scriptsize{Train}} & \makecell{\scriptsize{OpenDoor} \\ \scriptsize{Train}} & \makecell{\scriptsize{OpenDrawer}  \\ \scriptsize{Train}} & \makecell{\scriptsize{PushChair} \\ \scriptsize{Train}} & \makecell{\scriptsize{MoveBucket} \\ \scriptsize{Test}} & \makecell{\scriptsize{OpenDoor} \\ \scriptsize{Test}} & \makecell{\scriptsize{OpenDrawer} \\ \scriptsize{Test}} & \makecell{\scriptsize{PushChair} \\ \scriptsize{Test}} & \makecell{\scriptsize{Final} \\ \scriptsize{Score}}                   \\ \midrule
			
			Silver-Bullet-3D & \textbf{0.964} & \textbf{0.960} & \textbf{0.988} & \textbf{0.884} & \textbf{0.944} & \textbf{0.744} & \textbf{1.000} & \textbf{0.940} & \textbf{0.928} \\
			aidan-curtis     & 0.772          & 0.472          & 0.972          & 0.732          & 0.572          & 0.280          & 0.988          & 0.728          & 0.690          \\
			Fattonny         & 0.348          & 0.696          & 0.864          & 0.388          & 0.316          & 0.216          & 0.200          & 0.252          & 0.410          \\\bottomrule
		\end{tabular}\label{tab:no_restriction}
	\end{table}
	
	\section{Conclusion}
	
	In No Interaction Track of SAPIEN ManiSkill Challenge 2021, we integrate imitation learning with offline reinforcement learning, leading to high-quality learn-from-demonstrations solutions. Moreover, a three-stage network is designed to better exploit geometric and color information of observations. For No Restriction Track, we design a Heuristic Rule-based Method for object manipulation. Our future works include more in-depth studies of how imitation learning and offline reinforcement learning could be fused to boost policy learning and how to improve the performances with deep online reinforcement learning.
	
	\bibliography{gpl-iclr2022}
	\bibliographystyle{gpl-iclr2022}
	
\end{document}